# TgDLF2.0: Theory-guided deep-learning for electrical load forecasting via Transformer and transfer learning

Jiaxin Gao[1], Wenbo Hu[2], Dongxiao Zhang[3,4,*], and Yuntian Chen[1,*]

1 Eastern Institute for Advanced Study, Yongriver Institute of Technology, Zhejiang, P. R. China;

2 School of Computer and Information, Hefei University of Technology, Hefei, P. R. China;

3 Department of Mathematics and Theories, Peng Cheng Laboratory, Guangdong, P. R. China;

4 National Center for Applied Mathematics Shenzhen (NCAMS), Southern University of Science and Technology, Guangdong, P. R. China.

**ABSTRACT**

Electrical energy is essential in today's society. Accurate electrical load forecasting is beneficial for better scheduling of electricity generation and saving electrical energy. In this paper, we propose theory-guided deep-learning load forecasting 2.0 (TgDLF2.0) to solve this issue, which is an improved version of the theory-guided deep-learning framework for load forecasting via ensemble long short-term memory (TgDLF). TgDLF2.0 introduces the deep-learning model Transformer and transfer learning on the basis of dividing the electrical load into dimensionless trends and local fluctuations, which realizes the utilization of domain knowledge, captures the long-term dependency of the load series, and is more appropriate for realistic scenarios with scarce samples. Cross-validation experiments on different districts show that TgDLF2.0 is approximately 16% more accurate than TgDLF and saves more than half of the training time. TgDLF2.0 with 50% weather noise has the same accuracy as TgDLF without noise, which proves its robustness. We also preliminarily mine the interpretability of Transformer in TgDLF2.0, which may provide future potential for better theory guidance. Furthermore, experiments demonstrate that transfer learning can accelerate convergence of the model in half the number of training epochs and achieve better performance.

## 1. INTRODUCTION

Electrical energy is one of the most important forms of energy globally, with a significant impact on both industrial output and human life. Accurate short-term electrical load forecasting is becoming increasingly essential in order to effectively schedule electricity generation and save energy [1]. Indeed, a study shows that an improvement of only 1% in electricity load forecasting can save up to £10 million annually [2].

Researchers are increasingly interested in electrical load forecasting, and many forecasting models have been developed based on it. These models are separated into two categories: domain knowledge-based models and data-driven models.

For the knowledge-based models, Rahman and Bhatnagar created an expert system for load forecasting [1]. Although this system possesses good interpretability, it struggles to simulate complicated nonlinear interactions between features.

For the data-driven models, Park et al. used ANN to predict the future load [3]. However, their model is too simple, and it did not utilize temporal information to aid training. Bedi and Toshniwal used long short-term memory (LSTM) for electrical load forecasting [4, 5]. Jurasovic et al. utilized Transformer for day-ahead load forecasting and achieved good results [6, 7]. These data-driven models have high requirements on the amount of data, and without the assistance of domain knowledge, the models may be trapped in a local minimum and fail to achieve the highest accuracy during training.

Pure domain knowledge-based models and pure data-driven models are typically insufficient in handling complex problems and, as a consequence, some research has attempted to combine domain knowledge with data-driven models. Domain knowledge, on the other hand, is frequently employed mainly for feature engineering and has yet to be fully integrated with deep-learning algorithms [8, 9]. In this case, domain knowledge is often underutilized. A theory-guided framework was proposed to address this problem. Domain knowledge and data-driven algorithms can be fully integrated under this framework, which is based on the usage of first-principle models and empirical models as the reference and basis for model prediction. This framework has been successfully applied to many problems. For example, Wang et al. and He et al. used theory-guided neural networks (TgNNs) in the field of hydrology and achieved good results [10, 11]. Raissi et al. utilized the physics-informed neural network (PINN), which is essentially a type of theory-guided neural network (TgNN), for computational fluid dynamics and achieved good performance [12]. Karpatne et al. proposed theory-guided data science (TGDS), and presented five ways to integrate scientific knowledge and data science [13]. They also proposed the physics-guided neural network (PGNN) for lake temperature modeling [14]. He et al. developed the theory-guided full convolutional neural network (TgFCNN) to solve inverse problems in subsurface contaminant transport [15]. Li et al. proposed a TgNN as a prediction model for oil/water phase flow [16]. Chen et al. developed a kind of hard constraint model under the theory-guided framework to ensure that the model outputs obey known governing equations [17]. Previous theory-guided methods mostly used control equations, but in the field of electrical load forecasting, the physical process is too complicated, and there is still a lack of effective control equations. As a result, referring to the idea of theory-guided methods, the theory-guided deep-learning framework for load forecasting via ensemble long short-term memory (TgDLF) [18] was advanced.

TgDLF used domain knowledge to obtain the trend of the load, and employed the model EnLSTM [19, 20] to predict local fluctuations. TgDLF has achieved better results than pure knowledge-based models and pure data-driven models. In this work, we propose TgDLF2.0, an improved version of TgDLF. TgDLF2.0 replaces the EnLSTM model in TgDLF with a more expressive and efficient Transformer [6] model, which can improve accuracy and speed-up training.

Furthermore, due to the high cost of electrical load collection and data privacy, the problem of insufficient historical electrical load is frequently encountered in practice [21], which will affect the performance of models. In order to solve this problem, transfer learning [22-24] is used to aid training in TgDLF2.0, which can accelerate convergence of the model in half the number of training epochs and achieve better performance.

The contribution of this study is four-fold:
- (1) This study proposes a stronger and more robust model for short-term (only one day in the future) load forecasting.
- (2) This study reduces the time to train an electricity load forecasting model.
- (3) This study preliminarily mines interpretability of the deep-learning model for load forecasting.
- (4) This study solves the problem of insufficient historical load data.

## 2. METHODOLOGY

In this work, theory-guided deep-learning load forecasting 2.0 (TgDLF2.0) is used to predict the load ratio, and the real load can be recovered with the load ratio and historical load. In this section, the theory-guided framework is introduced first, followed by a demonstration of the deep-learning model Transformer and transfer learning. Finally, we show how to combine the theory-guided framework with Transformer and transfer learning for load forecasting.

### 2.1 Theory-guided framework

The theory-guided framework refers to using knowledge and theory to guide the training of deep-learning models. It combines the generality of human knowledge with the efficiency of data-driven models, as shown in Fig. 1. Research and experiments show that with the assistance of human knowledge, neural networks can avoid many detours to speed-up convergence, and often achieve higher accuracy. In this study, the theory-guided framework is used for load forecasting.

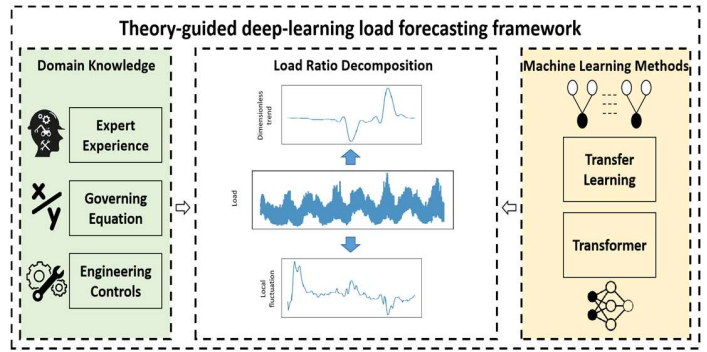

**Fig. 1.** Theory-guided framework.

Concretely, the electrical load data are a continuous series, and we can therefore convert the origin load series into the load ratio series. This conversion offers three main advantages: (1) it can convert non-stationary load series into stationary load ratio series; (2) it can compress the value range of features to a relatively small interval, which can speed-up convergence of the models; and (3) it can move the load data in different districts to a similar interval, which is similar to standardization, and can enhance the generalization ability of the models. This conversion is also crucial for transfer learning. The load ratio can then be further decomposed into the dimensionless trend and local fluctuation. The dimensionless trend can be obtained based on the physical mechanism and domain knowledge, and deep-learning models can focus more on predicting local fluctuation. Subsequent experiments prove that this decomposition can greatly improve prediction performance. This process can be expressed by Eq. (1):

$$\begin{aligned} L_{t+1} &= Ratio_{t+1} * L_t \\ &= f(x,t) * L_t \\ &= \big(f_1(x,t) + f_2(x,t)\big) * L_t \end{aligned}$$
$$f_1(x,t) = DT_{t+1}$$
$$f_2(x,t) = \delta_{t+1} \qquad (1)$$

where $L_{t+1}$ and $L_t$ represent the load at time *t+1* and *t* respectively; $Ratio_{t+1}$ represents the load ratio at time *t+1*; $f_1(x,t)$ is the dimensionless trend (*DT*); and $f_2(x,t)$ is the local fluctuation ($\delta$).

Furthermore, the theory-guided framework also emphasizes the importance of weather factors and calendar factors for electrical load forecasting [25, 26]. Under the theory-guided framework, instead of training many sub-models specifically to model weather and calendar factors [5], we utilize a unified model which fully considers a range of influencing elements, and is suitable for a variety of seasons and weather.

*2.2 Transformer*

The model Transformer has recently generated widespread interest among researchers, and it has been widely used in various machine-learning tasks, such as natural language processing (NLP), computer vision (CV), recommendation systems, mathematics [27-29], etc. Many promising applications are also based on Transformer, such as Bert, GPT-3, DALL-E, Codex [30-33], etc. Transformer can capture long-term dependency among features, and thus it is also proven to be effective in the field of time series processing [34].

Transformer is made up of two parts: an encoder and a decoder. The encoder is responsible for high-level feature extraction. It is a stack of encoder blocks, and each block contains a multi-head attention module and a position-wise feed-forward network (FFN). The decoder is utilized for the final prediction. It is a stack of decoder blocks, each of which contains two multi-head attention modules and an FFN. In addition, a residual connection is adopted for constructing a deeper model [6, 35].

Attention is the key component in Transformer, which allows features to interact with each other, and to produce more informative and expressive features. Attention comprises three important elements: query, key, and value. A query can be thought of as a questioning vector of a feature, which is used to ask other features about the interactions between them. A key and a value can be viewed as the index and value of a feature, respectively. The design of attention builds an interactive relationship among all of the features and can capture the long-term dependency (dependency information spanning up to 96 h, as input is 4 d of historical data) of the load ratio series, which is challenging for auto-regressive deep-learning models, such as recurrent neural networks (RNNs) and long short term memory (LSTM).

Specifically, an attention function is defined as a mapping function, which computes the inner product between the query and key vectors as the attention scores. The scaled dot-product attention used by Transformer is as follows:

$$\text{Attention}(Q,K,V) = \text{softmax}\left(\frac{QK^T}{\sqrt{D_K}}\right)V = AV \qquad (2)$$

where Q is queries; K is keys; V is values; Q, K, and V are generally obtained by applying some transformations to the original input, while the others are obtained through external input; $D_k$ represents the dimensions of keys; and A is often called an attention matrix, which can visualize the relationships between features. This attention matrix also provides interpretability for Transformer [36, 37], which constitutes another advantage over conventional models for time series data, such as LSTM.

Transformer usually adopts multi-head attention, instead of using a single attention function. The definition of multi-head attention is:

$$\text{MultiHeadAttn}(Q,K,V) = \text{Concat}(\text{head}_1, \ldots, \text{head}_H)W^O$$
$$\text{where head}_i = \text{Attention}(QW_i^Q, KW_i^K, VW_i^V) \qquad (3)$$

where $W_i^Q$, $W_i^K$, and $W_i^V$ are parameter matrices, projecting Q, K, and V to different representation subspaces, respectively. Studies have demonstrated that multi-heads can learn different dimensions of information [6, 38], so that the performance of Transformer that uses multi-head attention is usually superior.

Transformer can also better utilize the parallel computing power of GPUs [39,40], and thus it is usually more efficient than serial input models, such as LSTM.

*2.3 Transfer learning*

Transfer learning is a machine-learning method that takes a model developed on one task as an initial point and reuses it for developing a model on another task. The premise of transfer learning is that the goals of the two tasks are similar, and the data used by the two tasks are similar in terms of data structure. In our application, our task in each district is load forecasting, and the load data in different districts are similar in data structure, and thus transfer learning can be used between different districts.

In transfer learning, the load data of a district are chosen as the target data, and the load data of another district are chosen as the source data. The goal is to improve the performance of the deep-learning model in the target data. We first extract the dimensionless trend from the source data, and then train a model for the local fluctuation of the source data, which we refer to as the "source model". For transfer learning, the dimensionless trend from the source data is applied to the target data directly; in this way, the local fluctuation of the target data can be acquired. Then, the source model is applied to the local fluctuation of the target data. Three strategies are adopted for transfer learning while re-training the source model on the target data: (1) update all weights of the model while re-training; (2) fix the weights of the encoder, decoder, and embedding layers, and only update the weights of the last fully connected layer while re-training; and (3) fix the weights of the last fully connected layer, and only update the weights of the encoder, decoder, and embedding layers. These three strategies are illustrated in Fig. 2.

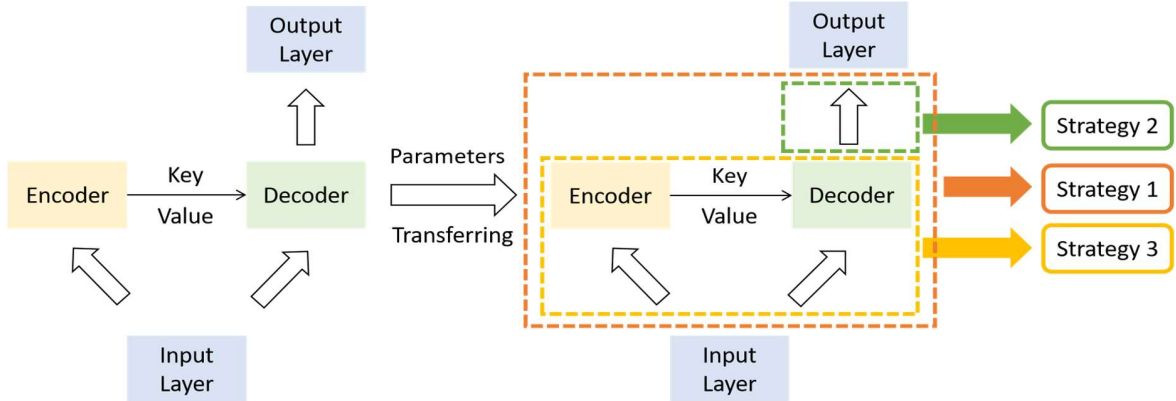

**Fig. 2.** Three strategies of transfer learning in TgDLF2.0.

*2.4 TgDLF2.0*

In this study, based on Transformer's expressive representation capability and the effectiveness of transfer learning, we propose theory-guided deep-learning load forecasting 2.0 (TgDLF2.0), which combines the theory-guided framework with Transformer and transfer learning.

In this study, we first extract the weekly average trend of the load ratio from the training data, a low pass filter is then applied to smoothing the trend, and the filtered trend is used as the dimensionless trend. Regarding local fluctuation, Transformer is used to predict it. The Transformer's input has three dimensions: (1) batch size, which is the number of samples used in each training iteration, and the batch size used in training is 512; (2) time step, which refers to how long the historical load is used for training, and the time step used is 96 h; and (3) the feature dimension, which contains a historical data feature together with four weather features and four date features, for a total of nine features. Transformer contains an encoder block and a decoder block, and the dimension of the hidden layer is 12. For the attention module, each attention is divided into two heads, and these two heads may learn different information while training. The attention module in Transformer is utilized to learn the interaction between all features at different time steps. The structure of the model and the data flow inside of the model are illustrated in Fig. 3.

We can acquire the load ratio by adding the dimensionless trend and local fluctuation together, and then the load ratio can be restored to the real load with the real load from the previous day, as shown in Eq. (1).

The process of TgDLF2.0 is shown in Fig. 4.

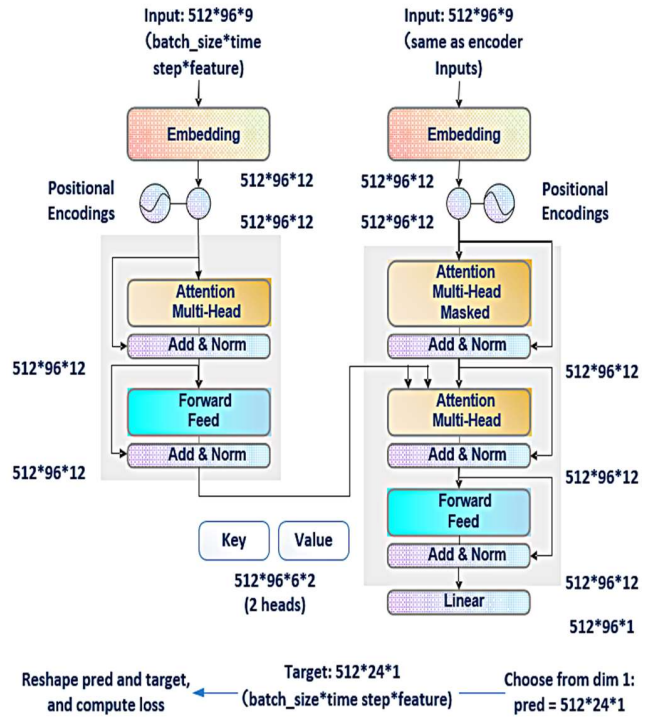

**Fig. 3.** Structure of Transformer.

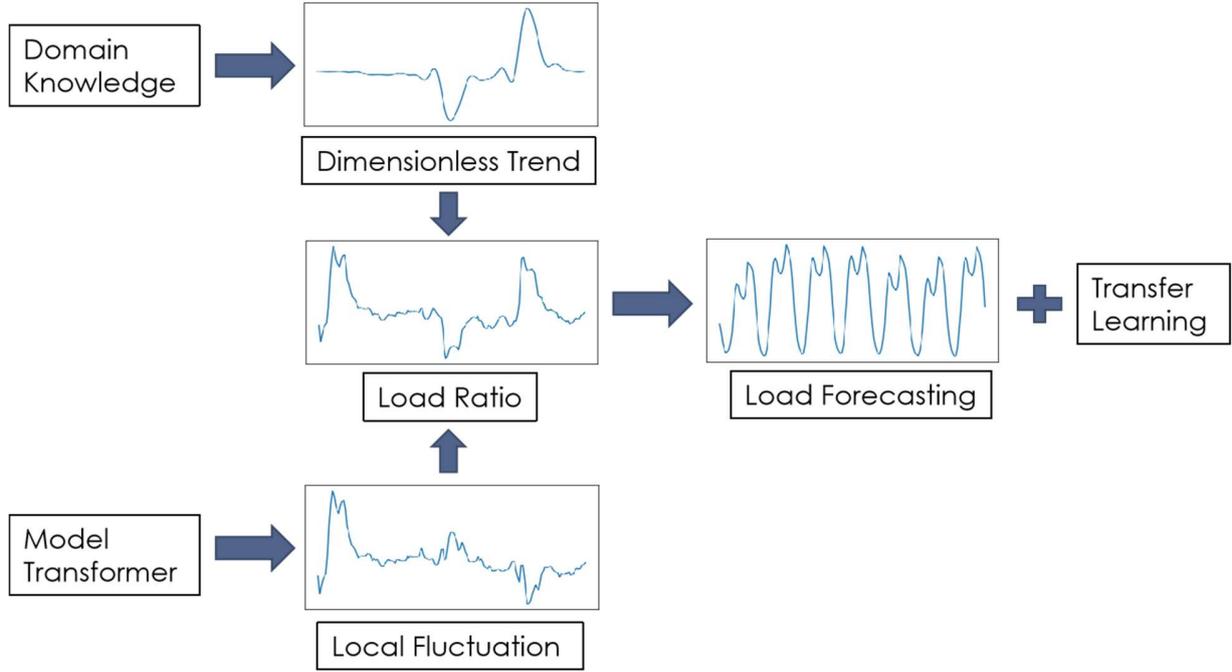

**Fig. 4.** Flowchart of TgDLF2.0.

### 3. EXPERIMENT

*3.1 Data description and experiment setting*

In this study, we take the electrical load data from 2008.01.01 to 2011.09.23 of 12 districts in Beijing, China, as a study case. The 12 districts are Chaoyang (CY), Haidian (HD), Fengtai (FT), Shijingshan (SJS), Pinggu (PG), Yizhuang (YZ), Changping (CP), Mentougou (MTG), Fangshan (FS), Daxing (DX), Miyun (MY), and Shunyi (SY). The location relationship of the 12 districts is illustrated in Fig. 5. These 12 districts can be divided into three groups according to correlation analysis: east Beijing (yellow); central Beijing (blue); and west Beijing (red). Since the load data and weather data are sampled per hour, there are 392,256 data in total.

The meteorological data in the 12 districts are also applied to load forecasting. The meteorological data include temperature, humidity, wind speed, and precipitation rate, which are four factors considered to be crucial for load forecasting.

Since both load data and weather data have a small proportion of missing values or outliers, preprocessing of the data is required. For the load data, two data preprocessing methods are mainly used. First, linear interpolation is used for missing value filling, as the effect of polynomial interpolation is not significantly better than linear according to the experiment. Second, the outliers are detected according to their neighborhood after filling the missing values. Since the load data are relatively stable in the short-term, points that are too different from their neighbors can be regarded as outliers, and thus we need to remove these outliers and use linear interpolation to fill in these values. For the meteorological data, the duplicated meteorological data are deleted at first, and missing values in precipitation rate are directly set to zero, taking expert experience into account.

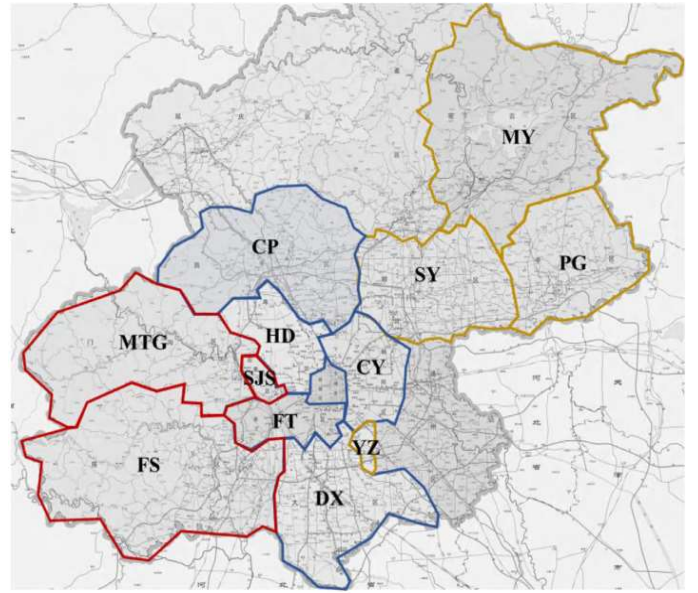

**Fig. 5.** Map of the 12 research districts in Beijing [18].

When using weather data, there are also two options: use historical weather data ($W_t$) as historical load data ($L_t$);

and use weather data for the next day ($W_{t+1}$). Since the real weather data for the next day cannot be obtained in practice, the weather data for the next day can be replaced by weather forecast data for the next day ($W'_{t+1}$). Weather forecast data are simulated with real weather data and normally distributed noise, as the weather forecast is always different from the actual weather. It is worth noting that the training set and test set contain real weather data, but they do not contain weather forecast data, so weather forecast data can be simulated in this way; when predicting the load for the next day in a real scenario, the real weather forecast data for the next day can be easily collected, and thus the real weather forecast data can be used. The formula for adding random noise to weather data is as follows:

$$W'_{t+1} = W_{t+1} * (1 + Noise * Proportion) \qquad (4)$$

where $W'_{t+1}$ and $W_{t+1}$ represent the weather forecast data and real weather data at time *t+1*, respectively; *Noise* represents normally distributed noise; and *Proportion* represents the proportion of added noise.

The results in the TgDLF show that the use of weather forecast data is significantly better than the use of historical weather data [18], which is logical. TgDLF2.0 also uses weather forecast data, and the proportion of added noise to real weather data is 5%.

In addition, we find that the load always changes greatly on Saturday and Monday, and thus two flags are added to the input to indicate whether the day is Monday or whether it is Saturday. Essentially, these two flags represent the switch of weekday and weekend. Moreover, it is well known that people's electricity consumption habits on weekends are not the same as those during weekdays, and people consume more electricity than usual in summer; consequently, two more flags are added to indicate whether the day is on a weekend and whether it is in the summer. Finally, the input has nine dimensions: one load ratio; four weather factors; and four calendar factors.

In TgDLF2.0, a moving window method is used to generate the samples for training and testing. Four days (4 d) of historical data are utilized to predict the load of 1 d in the future. The initial input window contains the first 4 d of historical load data, weather factors, and calendar factors, and the initial output window contains the load data for the fifth day. In this way, the first sample is generated. Each time the input window slides forward 1 d, the output window also slides forward 1 d, and more samples are generated. Each region can contribute 1,355 samples after sliding the load data, weather factors, and calendar factors.

### 3.2 Load forecasting experiments

In load forecasting experiments, to measure the performances of the models more objectively and make better use of the data, we adopt a four-fold cross-validation method [41], in which 12 districts are divided into four folds, and each experiment takes three folds as training data and one fold as test data. We conduct four experiments in total, so that the performances of the model in each district can be acquired.

**Table 1.** Prediction MSE of different models.

|  | ARMA | ARIMA | LSTM | DT | EnLSTM | TgDLF | Transformer | TgDLF2.0 |
|---|---|---|---|---|---|---|---|---|
| **PG** | 0.260 | 0.113 | 0.102 | 0.121 | 0.095 | 0.077 | 0.071 | **0.061** |
| **SJS** | 0.409 | 0.115 | 0.115 | 0.116 | 0.119 | 0.106 | 0.102 | **0.097** |
| **CY** | 0.071 | 0.062 | 0.052 | 0.065 | 0.046 | 0.032 | 0.027 | **0.023** |
| **YZ** | 0.160 | 0.121 | 0.091 | 0.126 | 0.089 | 0.080 | 0.068 | **0.056** |
| **MTG** | 2.296 | 0.324 | 0.111 | 0.123 | 0.120 | 0.107 | 0.105 | **0.096** |
| **FT** | 0.097 | 0.070 | 0.058 | 0.064 | 0.053 | 0.030 | 0.029 | **0.028** |
| **MY** | 0.188 | 0.083 | 0.072 | 0.078 | 0.065 | 0.049 | 0.050 | **0.044** |
| **FS** | 0.363 | 0.102 | 0.093 | 0.096 | 0.091 | 0.075 | 0.079 | **0.071** |
| **CP** | 0.144 | 0.069 | 0.059 | 0.056 | 0.053 | 0.040 | 0.038 | **0.035** |
| **SY** | 0.174 | 0.091 | 0.078 | 0.100 | 0.070 | 0.055 | 0.048 | **0.041** |
| **HD** | 0.081 | 0.078 | 0.064 | 0.081 | 0.053 | 0.042 | 0.038 | **0.031** |
| **DX** | 0.151 | 0.054 | 0.059 | 0.062 | 0.051 | 0.035 | 0.033 | **0.028** |
| **AVG** | 0.366 | 0.107 | 0.079 | 0.091 | 0.075 | 0.061 | 0.057 | **0.051** |

In experiments, mean square error (MSE) metrics are used to evaluate models. The results of cross-validation are shown in Table 1. Four baselines are selected: ARMA and ARIMA [42, 43] are the baselines as traditional statistical models; LSTM is the baseline as a classical time-series deep-learning model; and the baseline DT represents a model that predicts the load only based on the dimensionless trend, and does not utilize machine-learning models to predict local fluctuations. In addition, we list the results of TgDLF, which is the previous state-of-the-art (SOTA) model. To demonstrate the importance of domain knowledge, the results of TgDLF and TgDLF2.0 without the assistance of dimensionless trend (EnLSTM and Transformer, respectively) are also listed in Table 1.

It is shown in Table 1 that TgDLF2.0 has an obvious advantage in each district. The MSE of TgDLF2.0 is 86.1%, 52.3%, and 35.4% lower than the MSE of ARMA, ARIMA, and LSTM, respectively. In addition, the MSE of TgDLF2.0 is 44.0% lower than the MSE of DT, which proves that only using dimensionless trends is insufficient. Moreover, TgDLF2.0 surpasses its previous version TgDLF with 16.3% in MSE, and due to the training efficiency of Transformer, TgDLF2.0 saves more than half of the training time compared to TgDLF, as shown in Table 2.

**Table 2.** Training time of TgDLF and TgDLF2.0.

|  | AVG | Fold 0 | Fold 1 | Fold 2 | Fold 3 |
|---|---|---|---|---|---|
| **TgDLF** | 769 s | 762 s | 758 s | 774 s | 780 s |
| **TgDLF2.0** | 344 s | 348 s | 328 s | 355 s | 345 s |

Furthermore, TgDLF is superior to EnLSTM with 23.0%, and TgDLF2.0 is superior to Transformer with 10.5%. These results demonstrate the importance of domain knowledge.

### 3.3 Additional analysis of load forecasting experiments

In this subsection, we first visually show the prediction performance of TgDLF2.0, and then analyze the interpretability of Transformer in TgDLF2.0. Finally, we test the robustness of TgDLF2.0 against weather noise [44].

The predicted outcomes of TgDLF2.0 in the Fengtai district are used as an example in Fig. 6 to demonstrate the model performance in further detail. The electrical load (including local fluctuation, dimensionless trend, load ratio, and real load) is represented by the ordinate, and time is shown by the abscissa. We choose the time period from 2008.08.08 to 2008.09.17, during which the Olympic and Paralympic Games were held in Beijing. The red lines represent the prediction results of TgDLF2.0; whereas, the black lines represent the actual value. It can be seen that the prediction of the local fluctuation by the model Transformer is good in most cases, but in some places with large changes, the prediction still has a certain bias. However, with the addition of human knowledge (dimensionless trend), this bias is reduced. We finally restore the load ratio back to the real load. It can be seen that the predicted load and the actual load are consistent. Fig. 6 intuitively shows the excellent prediction performance of TgDLF2.0.

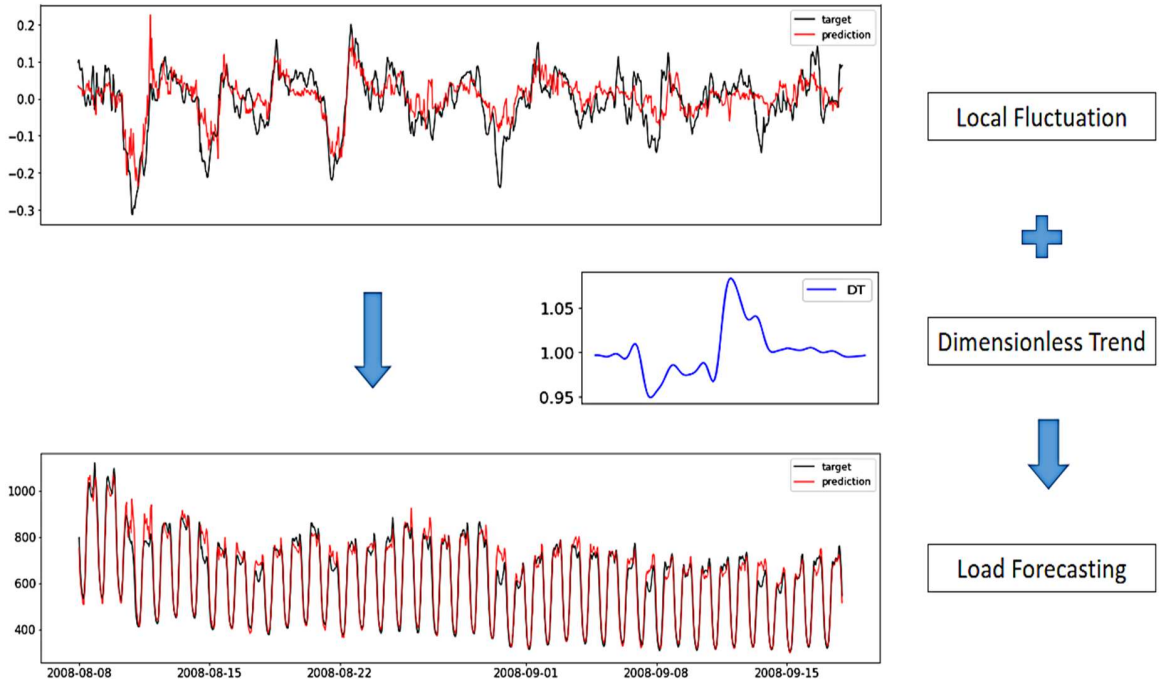

**Fig. 6.** Load forecasting of TgDLF2.0 in the Fengtai district.

The Transformer in TgDLF2.0 not only possesses high expressiveness, but also has good interpretability. The interpretability of Transformer is mainly reflected through attention [36, 37], and it is possible to observe what Transformer has learned by visualizing the attention matrix.

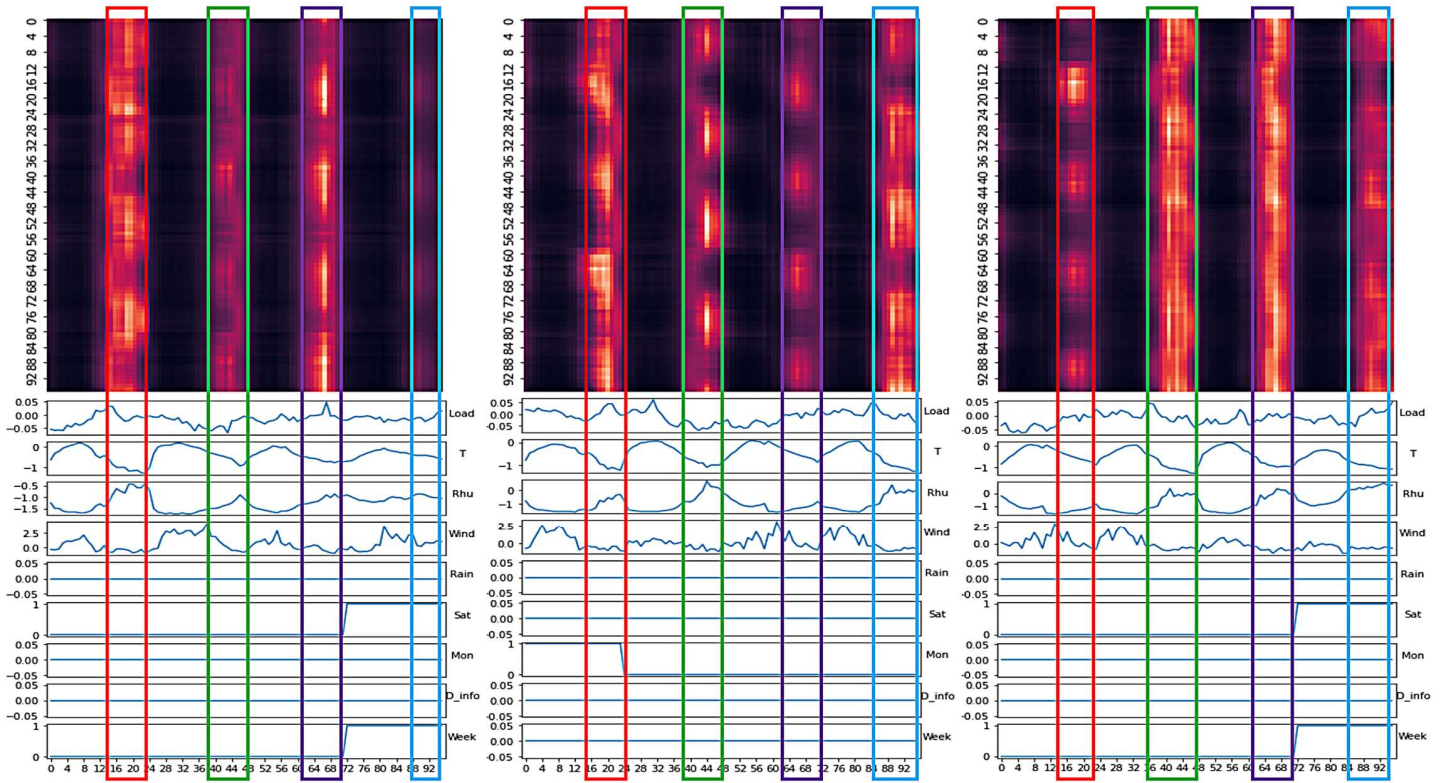

**Fig. 7.** Several typical attention matrices of Transformer.

Several typical attention matrices of Transformer and their corresponding samples are presented in Fig. 7. The three figures above are attention matrices, and the length of the horizontal and vertical coordinates of each attention matrix is 96, representing the 96 h of input. Each point in the attention matrix can be understood as a combined feature point formed by the interaction of the original features: for example, the point (15, 20) can be understood as the combined feature point formed by the interaction of the original feature of the 15$^{th}$ hour and the 20$^{th}$ hour. Overall, the larger is the value of this point, the greater is the influence that it has on the final prediction of the model Transformer. In the figure of attention matrix, the depth of the color can be used to represent the value of the point. Overall, the brighter is the color, the larger is the value in the attention matrix, and it can also be proven that the combined feature represented by the point is more important for the prediction of the model. The three figures below are samples corresponding to the attention matrices. The abscissa of the figure also represents the input of 96 h. The figure can be divided into nine subfigures, with each subfigure representing a feature of the input (load, temperature, humidity, wind speed, precipitation rate, whether it is Saturday, whether it is Monday, calendar effect, and whether it is a weekend, from top to bottom), and the ordinate of each subfigure represents the value of the feature. These three attention matrices are band-shaped, and the highlighted parts are concentrated from 3:00 pm to 11:00 pm every day. In fact, we found that the attention matrices corresponding to most samples are band-shaped, and the highlighted time periods are basically the same. Since the local fluctuations to be predicted represent uncertainty, we suspect that this is due to the higher uncertainty of people's activity between 3:00 pm and 11:00 pm (people are generally at work before 3:00 pm and resting after 11:00 pm, and the uncertainty of people's activities within these two time periods will be much smaller). Since the time period from 3:00 pm to 11:00 pm can provide more information for the model prediction, the model will pay more attention to this time period of the day. In addition to the band-shaped attention matrices, some attention matrices are checkerboard-shaped, and there is a small number of bright spot-shaped attention matrices. Example attention matrices and corresponding samples are provided in Appendix A. We believe that the samples corresponding to these attention matrices contain very complex nonlinear mapping relationships, and we will analyze them in the future. It can be seen that the Transformer model has achieved good prediction results based on these outstanding understandings of the samples. The interpretability of Transformer not only shows how the model works, but also provides future potential for better theory guidance.

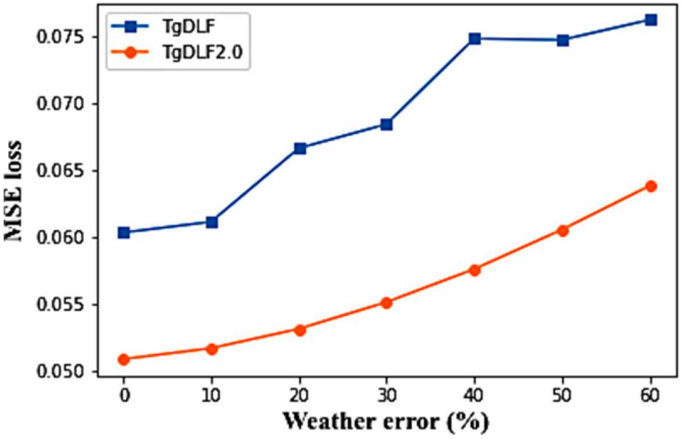

**Fig. 8.** The MSE loss of TgDLF and TgDLF2.0 with different scales of noise in the weather forecast data.

It is well known that the model's prediction accuracy can be significantly increased by using accurate weather forecast data [26]. Since weather forecast data cannot be absolutely accurate, we also conduct an experiment to assess the effect of noisy weather forecast data on load forecasting accuracy. In order to simulate scenarios with varying weather forecast accuracy, normally distributed random errors with standard deviations of 10%, 20%, 30%, 40%, 50%, and 60%, respectively, are added to the weather data of the test set while testing.

The experimental results are displayed in Fig. 8. The normally-distributed error added to the weather forecast data appears on the x-axis, while the MSE loss appears on the y-axis. As the amount of noise in the weather forecast data grows, it is evident from Fig. 8 that both TgDLF and TgDLF2.0's predictive capabilities decline as predicted, but TgDLF2.0's rate of decline is slower, demonstrating that it is comparatively more robust to weather noise. Furthermore, even if the proportion of weather noise reaches 50%, the performance of TgDLF2.0 is still the same as that of TgDLF without weather noise, which further reflects the high accuracy and robustness of TgDLF2.0. Detailed forecast results for different districts are given in Appendix B.

*3.4 Transfer learning experiments*

In transfer learning experiments, we choose the load data of a district as the target data and the load data of another district as the source data. The goal is to improve the performance of the model in the target data. In transfer learning, the source data are often sufficient, while the target data are not sufficient. Due to the difference in the distribution of source data and target data, it is difficult to directly combine the source data and the target data to train the deep-learning model. To match the transfer learning scenario [22-24], we only select 192 d of load data in the source district, and then select separately 128, 64, 32, and 16 d of load data in the target district for training. We also select 64 d of load data in the target district for testing. Noise is also added to weather data to test the robustness of the model.

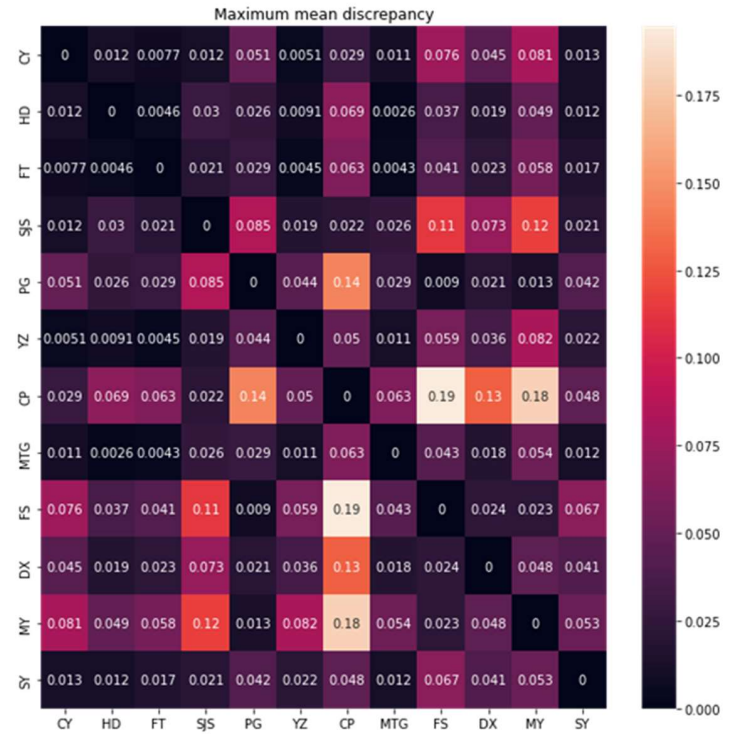

**Fig. 9.** The similarity between loads of different districts.

The maximum mean discrepancy (MMD) algorithm [45, 46] can be used to measure the similarity between load data in different districts. Fig. 9 shows the MMD distance of the load data in 12 districts. Overall, the smaller is the value, the higher is the similarity of the load data of the two districts. In each group, the load data with high similarity are selected for the transfer learning experiment. We finally choose three sets of districts (MTG, FS), (MY, PG), and (CY, FT) (the first district in a set is the district of the source data, and the latter is the district of the target data), and transfer learning achieved good results on all three sets of load data. As an example, the set (MTG, FS) is discussed next.

To ensure the objectivity of the experiment, we conduct five independent repeated experiments, and the parameters of the model are randomly initialized in each experiment. We can observe how the initial model performs on the target test set with and without transfer learning.

It is shown in Table 3 that the initial loss of the model is high when transfer learning is not used for initialization, and the loss caused by different initialization parameters is markedly different. After using transfer learning, the target data can be optimized from a very low loss, and it only needs half of the training epochs to converge compared to no transfer learning. Transfer learning can provide better initialization, which not only makes convergence faster, but also makes training more stable (the initial losses of the model on target data are similar). The loss curve is provided in Appendix C.

The final performances of the model under different transfer strategies, different scales of weather noise, and different samples of target load data are also illustrated in Fig. 10.

**Table 3.** Prediction MSE (load ratio) of the initial model with and without transfer learning.

|  | AVG | Exp 0 | Exp 1 | Exp 2 | Exp 3 | Exp 4 |
|---|---|---|---|---|---|---|
| No Transfer | 0.190 | 0.100 | 0.078 | 0.167 | 0.331 | 0.275 |
| Transfer | 0.008 | 0.007 | 0.008 | 0.008 | 0.007 | 0.008 |

Transfer learning on the set (MTG, FS) outperforms no transfer learning, as shown in Fig. 10. The added weather noise is normally distributed noise with a mean of zero, and the variance of the weather noise is represented by the abscissa in Fig. 10. As the variance of weather noise increases, the performance of TgDLF2.0 does not decrease obviously, which proves that TgDLF2.0 possesses strong robustness. As the target training data decrease, the performance of the model will decrease, but the effect of transfer learning is still significantly better than no transfer learning. The first transfer learning strategy (i.e., update all weights of the model while re-training) is relatively the best among the three strategies for the set (MTG, FS). The transfer learning results of the set (MY, PG) and the set (CY, FT) are provided in Appendix C.

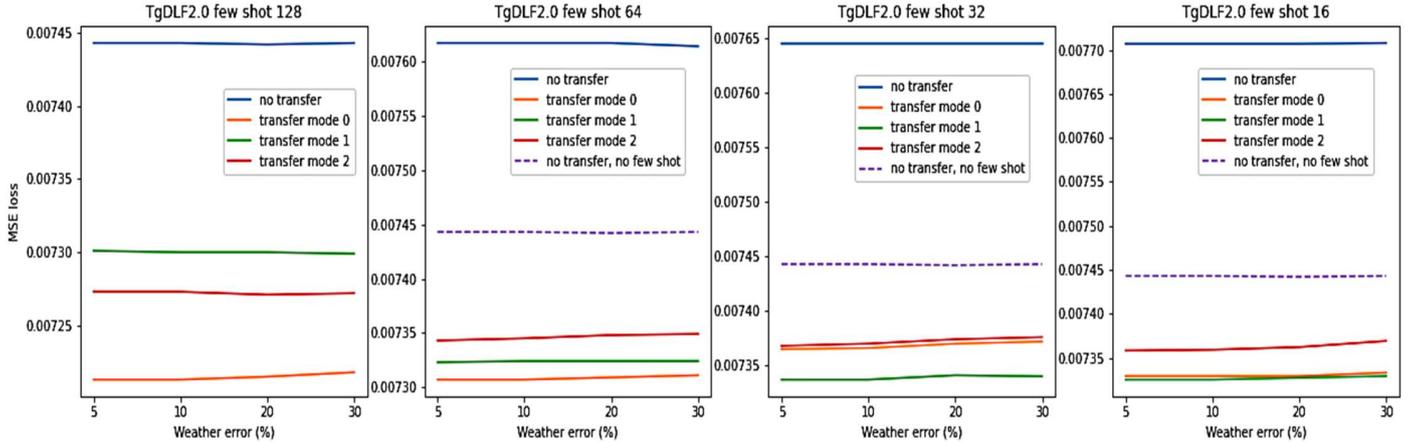

**Fig. 10.** Transfer learning results of the set (MTG, FS).

### 4. DISCUSSION

In this study, we proposed TgDLF2.0, a more accurate and efficient model for short-term electrical load forecasting. In TgDLF2.0, the dimensionless trends are obtained from expert experience or historical data, and the local fluctuations are predicted by the model Transformer based on historical load data, weather factors, and calendar factors. TgDLF2.0 produces good results, and makes the following contributions: (1) TgDLF2.0 obtains a 16% increase in accuracy over the previous version TgDLF. TgDLF2.0 also has the same accuracy as TgDLF without noise, even when there is 50% weather noise, demonstrating its robustness; (2) when compared to TgDLF, TgDLF2.0 requires 50% less training time; (3) the Transformer in TgDLF2.0 offers some interpretability, which may provide better theory guidance; and (4) TgDLF2.0 incorporates transfer learning to address the issue of limited data, which can accelerate convergence of the model in half the number of training epochs and achieve better performance. Furthermore, TgDLF2.0 can also be effective in load forecasting when deployed to new regions with scant historical data with the aid of transfer learning.

To the best of the authors' knowledge, TgDLF2.0 is the first applicable solution that combines the model Transformer with human knowledge, and mines the interpretability of Transformer, for short-term electrical load forecasting. Furthermore, it is also the first to use transfer learning to solve the problem of insufficient historical electrical load data. This work is innovative and may contribute meaningfully to the work of others.

We consider the following as future work for TgDLF2.0: (1) the current dimensionless trend is a simple filtered weekly average trend, and it is possible to choose a more complicated and accurate dimensionless trend; (2) the current loss function for model training is the ordinary MSE loss function, which only considers the numerical difference between the predicted value and the actual value. It is possible to further add human knowledge to the loss function in the form of regularization terms [12, 47]; (3) the interpretabilities of checkerboard-shaped attention matrices and bright spot-shaped attention matrices has shown a distinct structure pattern for the learned deep-learning model, which provides future potential for better theory guidance; and (4) the current attention matrix of Transformer is utilized to learn the interaction between all features at different time steps. It is possible to construct a 3D attention matrix [48] to learn the interaction between different features at different time steps. If one of these proposals is proven to work, TgDLF2.0 will achieve better performance.

**ACKNOWLEDGMENT**

This work is funded by the National Natural Science Foundation of China (Grant No. 62106116), the Shenzhen Key Laboratory of Natural Gas Hydrates (Grant No. ZDSYS20200421111201738), and the SUSTech - Qingdao New Energy Technology Research Institute.

**APPENDIX A. ADDITIONAL ATTENTION MATRICES OF TRANSFORMER**

A checkerboard-shaped attention matrix and its corresponding sample are shown in Fig. A.1. In the attention matrix of this pattern, bright and dark spots alternate regularly. A bright spot-shaped attention matrix and its corresponding sample are presented in Fig. A.2. In the attention matrix of this pattern, only a small part are bright spots, and the rest are all dark spots.

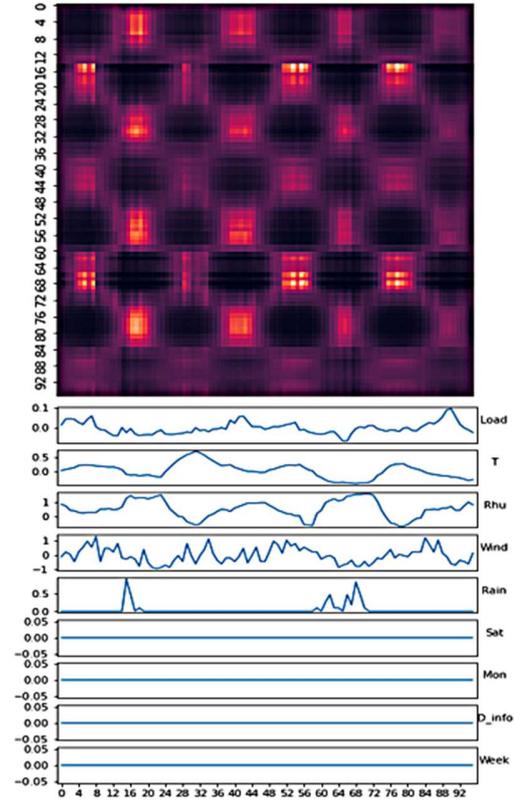

**Fig. A.1.** A checkerboard-shaped attention matrix and its corresponding sample.

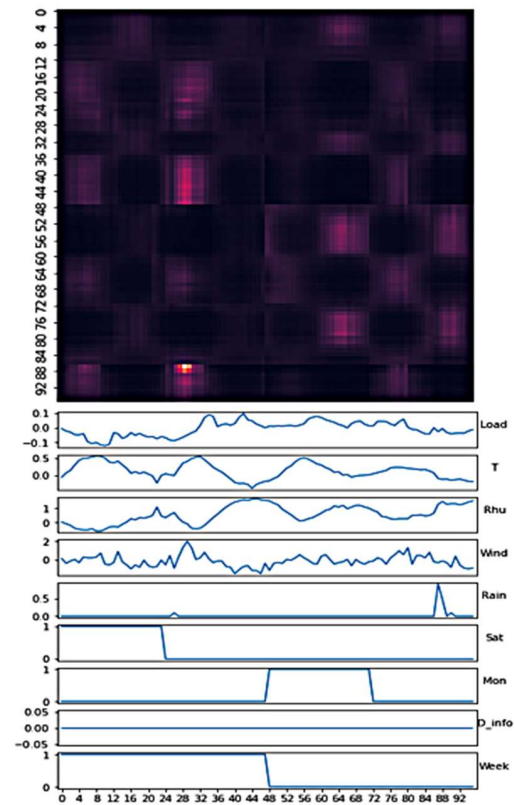

**Fig. A.2.** A bright spot-shaped attention matrix and its corresponding sample.

## APPENDIX B. DETAILED FORECAST EFFECTS

Detailed forecast effects of TgDLF2.0 with different scales of noise in the weather forecast data for different districts are listed in Table B.1. The header of the table represents the scale of normally-distributed error added to the weather forecast data.

**Table B.1.** The MSE loss of TgDLF2.0 with different scales of noise in the weather forecast data.

|     | 0     | 10%   | 20%   | 30%   | 40%   | 50%   | 60%   |
|-----|-------|-------|-------|-------|-------|-------|-------|
| PG  | 0.061 | 0.062 | 0.064 | 0.066 | 0.070 | 0.074 | 0.078 |
| SJS | 0.097 | 0.097 | 0.098 | 0.099 | 0.100 | 0.101 | 0.102 |
| CY  | 0.023 | 0.023 | 0.024 | 0.026 | 0.028 | 0.030 | 0.033 |
| YZ  | 0.056 | 0.059 | 0.062 | 0.066 | 0.070 | 0.073 | 0.077 |
| MTG | 0.096 | 0.097 | 0.098 | 0.099 | 0.100 | 0.101 | 0.102 |
| FT  | 0.028 | 0.028 | 0.030 | 0.033 | 0.037 | 0.040 | 0.044 |
| MY  | 0.044 | 0.046 | 0.047 | 0.049 | 0.052 | 0.055 | 0.057 |
| FS  | 0.071 | 0.072 | 0.073 | 0.074 | 0.076 | 0.077 | 0.078 |
| CP  | 0.035 | 0.035 | 0.037 | 0.040 | 0.042 | 0.045 | 0.049 |
| SY  | 0.041 | 0.042 | 0.043 | 0.045 | 0.049 | 0.053 | 0.058 |
| HD  | 0.031 | 0.031 | 0.031 | 0.032 | 0.033 | 0.033 | 0.034 |
| DX  | 0.028 | 0.028 | 0.029 | 0.032 | 0.036 | 0.043 | 0.052 |
| AVG | 0.051 | 0.052 | 0.053 | 0.055 | 0.058 | 0.061 | 0.064 |

## APPENDIX C. ADDITIONAL RESULTS OF TRANSFER LEARNING

The loss curve of the set (MTG, FS) is shown in Fig. C.1. It can be seen that the target data can be optimized from a very low loss with transfer learning, and it only needs half of the training epochs to converge compared to no transfer learning. The transfer learning results of the set (MY, PG) and the set (CY, FT) are shown in Fig. C.2 and Fig. C.3, respectively. The performances of transfer learning on these two sets are also significantly better than those of no transfer learning. The third transfer learning strategy (i.e., update the weights of the encoder, decoder, and embedding layers of the model while re-training) is relatively the best among the three strategies for the set (MY, PG), and the second transfer learning strategy (i.e., update the weights of the last fully connected layer of the model while re-training) is relatively the best for the set (CY, FT). Overall, the effects of the three strategies are not substantially different.

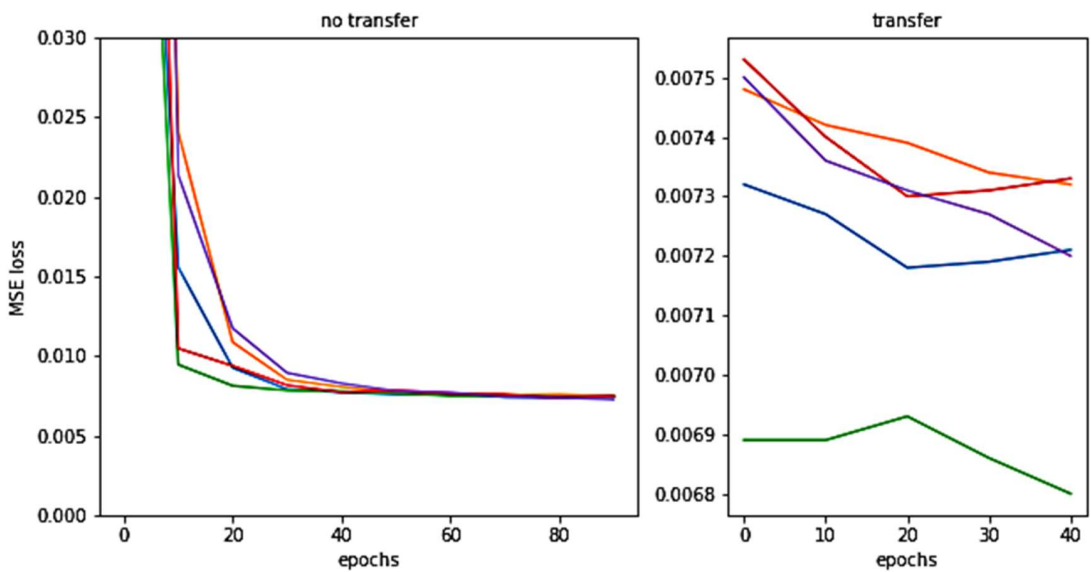

**Fig. C.1.** Loss curve of the set (MTG, FS).

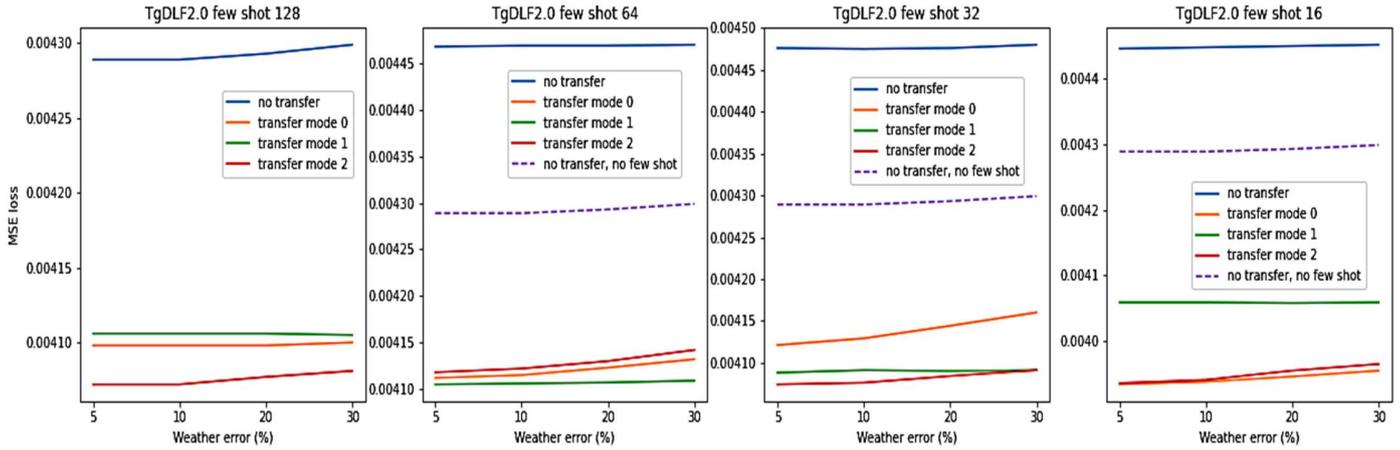

**Fig. C.2.** Transfer learning results of the set (MY, PG).

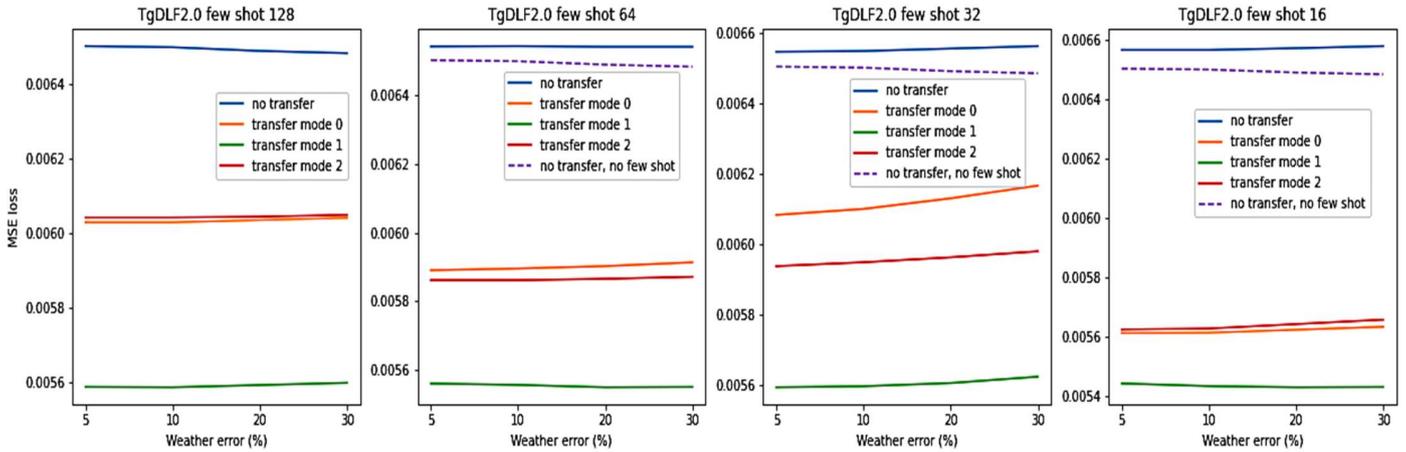

**Fig. C.3.** Transfer learning results of the set (CY, FT).